\documentclass[runningheads]{llncs}
\usepackage{graphicx}
\usepackage{amsmath,amssymb} 
\usepackage{color}
\newcommand{\Lagr}{\mathcal{L}}
\usepackage{natbib}
\begin{document}

\title{Perceptual Conditional Generative Adversarial Networks for End-to-End Image Colourization} 
\titlerunning{CuPGAN} 


\author{Shirsendu Sukanta Halder\inst{1} \and
Kanjar De\inst{1} \and
Partha Pratim Roy\inst{1}}


%

\authorrunning{Halder et al.} 


\institute{Indian Institute of Technology Roorkee, Roorkee-247667, India
\email{\{shalder@cs.iitr.ac.in, \{kanjar.cspdf2017, proy.fcs\}@iitr.ac.in\}}}

\maketitle

\begin{abstract}
Colours are everywhere. They embody a significant part of human visual perception. In this paper, we explore the paradigm of hallucinating colours from a given gray-scale image. The problem of colourization has been dealt in previous literature but mostly in a supervised manner involving user-interference. With the emergence of Deep Learning methods numerous tasks related to computer vision and pattern recognition have been automatized and carried in an end-to-end fashion due to the availability of large data-sets and high-power computing systems. We investigate and build upon the recent success of Conditional Generative Adversarial Networks (cGANs) for Image-to-Image translations. In addition to using the training scheme in the basic cGAN, we propose an encoder-decoder generator network which utilizes the class-specific cross-entropy loss as well as the perceptual loss in addition to the original objective function of cGAN. We train our model on a large-scale dataset and present illustrative qualitative and quantitative analysis of our results.  Our results vividly display the versatility and the proficiency of our methods through life-like colourization outcomes.
\end{abstract}

\section{Introduction}
\label{sec:intro}
\begin{figure}[!htbp]
 \centering
 \begin{tabular}{cc}
  \includegraphics[width=0.3\linewidth]{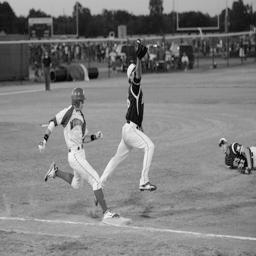}&
  \includegraphics[width=0.3\linewidth]{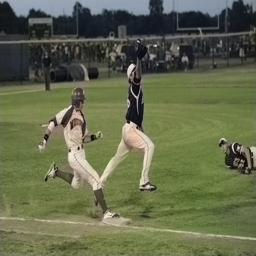}
 \end{tabular}
\caption{The process of image colourization focuses on hallucinating realistic colours.}
  \label{fig:firstfigure}
\end{figure}

Colours enhance the information as well as the expressiveness of an image. Colour images contain more visual information than a gray-scale image and is useful for extracting information for high-level tasks. Humans have the ability to manually fill gray-scale images with colours taking into consideration the contextual cues. This clearly indicates that black-and-white images contain some latent information sufficient for the task of colourization. The modelling of this latent information to generate chrominance values of pixels of a target gray-scale image is called colourization. Image colourization is a daunting problem because a colour image consists of multi-dimensional data according to defined colour-spaces whereas a gray-scale image is just single-dimensional. The main obstacle is that different colour compositions can lead to a single gray level but the reverse is not true. For example, multiple shades of blue are possible for the sky, leaves of trees attain colours according to seasons,  different colours of a pocket billiards (pool) ball. The aim of the colourization process is not to hallucinate the exact colour of an object (See Fig. \ref{fig:firstfigure}), but to transfer colours plausible enough to fool the human mind. Image colourization is  a widely used technique in commercial applications and a hot researched topic in the academia world due to its application in heritage preservation, image stylization and image processing. 

In the last few years, Convolutional Neural Networks (CNNs) have emerged as compelling new state-of-the-art learning frameworks for Computer Vision and Pattern Recognition \cite{deep1}; \cite{deep2} applications. With the recent advent of Generative Adversarial Networks (GANs) \cite{gans2014}, the problem of transferring colours have also been explored in the context of GANs using Deep Convolutional Neural Networks \cite{isola2017}; \cite{nazeri2018image}. The proposed method involves utilizing conditional Generative Adversarial Networks (cGANs) modelled as an image-to-image translation framework.\\ 
\indent The main contribution of this paper is proposing a variant of Conditional-GANs which tries to learn a functional mapping from input grayscale image to output colourized image by minimizing the adversarial loss, per pixel loss, classification loss and the high-level perceptual loss. The proposed model is validated using an elaborate qualitative and quantitative comparison with existing methods. A detailed ablation study which demonstrates the efficiency and potential of our model over baselines is also presented.


The rest of this paper is arranged as Section \ref{sec:relwork} describes previous works described in literature. Section \ref{sec:proposedmethod} describes our proposed Perceptual-cGAN. Section \ref{sec:results} demonstrates a detailed qualitative and quantitative analysis of images colourized by our proposed system along with ablation studies of different objective functions. The final Section \ref{sec:conclusion} consists of concluding remarks.

\section{Related Work}
\label{sec:relwork}

The problem of colourization has been studied extensively by numerous researchers due to its importance in real-world applications. Methods involving user-assisted scribble inputs were the primary methods explored for the daunting problem of assigning colours to a one-dimensional gray-scale image. \cite{levin2004} used user-based scribble annotations for video-colourization which propagated colours in the spatial and temporal dimensions ensuring consistency to produce colourized films. This method relied on neighbouring smoothness by optimization through a quadratic cost function. \cite{shapiro2005} proposed a framework where the geometry and the structure of the luminance input was considered on top of user-assisted information for colourization by solving a partial differential equation. \cite{noda2006} formulates the colourization problem into a maximum a posteriori (MAP) framework using Markov Random Field (MRF) as a prior. \\
\indent User-involved methods achieve satisfying results but they are severely time-consuming, needs manual labour and the efficacy is dependant on the accuracy of the user interaction. In order to alleviate these problems, researchers resorted to example-based or reference-based image colourization. Example-based methods require a reference image from the user for the transfer of colours from a source image to a target gray-scale image.\cite{reinhard2001} used simple statistical analysis to establish mapping functions that impose a source image's colour characteristics onto a target image. \cite{welsh2002} proposed to transfer only chrominance values to a target image from a source by matching the luminance and textural information. The luminance values of the target image are kept intact. The success of example-based methods rely heavily on the selection of an appropriate user-determined reference image. Hence it faces a limitation as there is no standard criterion for choosing an example image and hence depends on the user skill. Also the reference image may have different illumination conditions resulting in anomalous colour transfer. \\
\indent In recent years, the use of Deep Learning methods have emerged as prominent methods for the task of Pattern Recognition, even outperforming human ability \cite{he2015}. They have shown promising results in colourization by directly mapping gray-scale target images to output colors by learning to combine low level features and high-level cues. \cite{cheng2015deep} proposed the first deep learning framework using low-level features from patch, intermediate DAISY features \cite{tola2010daisy} and a high-level semantic features as output to a neural network ensemble. \cite{zhang2016colorful} proposed an automatic colourization process by posing it as a multinomial classification task for \textit{ab} values. They observed that in natural images the distribution of \textit{ab} values were desaturated and used class re-balancing for obtaining accurate chromatic components. A joint end-to-end method using two parallel CNNs and incorporation of local and global priors was proposed by \cite{iizuka2016}. The potential of GANs in learning expressive features trained under an unsupervised scheme propelled researchers to use GANs for the task of colourization. \cite{isola2017} explored the idea of Conditional-GANs in an image-to-image framework by learning a loss function for mapping input to output images.
\section{Proposed Method}
\label{sec:proposedmethod}

In this section, we describe our proposed algorithm along with the architecture of the Convolutional Neural Networks employed for our method. We explore the situation of GANs in a conditional setting \cite{isola2017}; \cite{nazeri2018image} where the objective function of the generator is constrained on the conditional information and consequentially generates the output. The addition of this conditional variables enhanced the stability of the original GAN frame-work proposed by \cite{gans2014}. Our proposed method builds upon the cGAN by incorporating adjunct losses. In addition to the adversarial loss and the per-pixel constraint, we add the perceptual loss and the classification loss in our objective function. We explain in detail about the loss functions and the networks used. 

\subsection{Loss function}
\label{sub:objectivegan}

Conditional-GANs \cite{isola2017} try to learn a mapping from the input image $I$ to output $J$. In this case, the generator not only aspires to win the mini-max game by fooling the discriminator but also has to be as close as possible with the ground truth. The per-pixel loss is utilized for this constraint. While the discriminator network of the cGAN remains the same when compared to the original GAN, the objective function of the Generator networks are different. The objective function of cGAN \cite{isola2017} is

\begin{equation}
\label{eq:isolagan}
     \Lagr_{cGAN} = min_G max_D \mathop{\mathbb{E_{I}}} [\log(1 - D(G(I)))] + \mathop{\mathbb{E_{I,J}}}[\log D(J)]
\end{equation}

Here $I$ represents the input image, $J$ represents the coloured image, $G$ represents the Generator amd $D$ represents the Discriminator. Isola \textit{et al.} \cite{isola2017} established through their work that $\Lagr_1$ is better due to their less blurring effect and also helps in reducing artifacts that are introduced by using only cGAN. Our proposed model Colourization using Perceptual Generator Adversarial Network (CuPGAN), builds up on this cGAN with incorporated additive perceptual loss and classification loss. Traditional loss functions that operate at per-pixel level are found to be limited in their attempt to capture the contextual and the perceptual features. Recent researches have demonstrated the competence of loss functions based on the difference of high-level feature in generating compelling visual performance [ \cite{perceptual2016}]. However, in many cases they fail to preserve the low-level colour and texture information \cite{perceptualderaining2017}. Therefore, we incorporate both the perceptual loss and the per-pixel loss for preservation of both the high-level and the low-level features. The high-level features are extracted from VGGNet model by \cite{Simonyan14c} trained on the ImageNet dataset \cite{imagenet2009}. Like \cite{iizuka2016}, we add an additional classification loss that helps our model to guide the training 
of the high-level features. 

For input gray-scale image $I$ and colourized image $J$, the perceptual loss between $I$ and $J$ in this case is defined as:

\begin{equation}
\label{eq:percepequation}
    \Lagr_{per} = \frac{1}{C,H,W} \sum_{c=1}^{C} \sum_{w=1}^{W} \sum_{h=1}^{H} \lVert V_i(G(I^{c,w,h})) - V_i(J^{c,w,h}) \rVert_2
\end{equation}

where $V_i$ represents a non-linear transformation by the $i^{th}$ layer of the VGGNet, $G$ represents the transformation function of the generator and $(C, W, H)$ are the channels, width and height respectively. The $\Lagr_{L1}$ loss between $I$ and $J$ is defined as:

\begin{equation}
\label{eq:perpixelloss}
    \Lagr_{L1} = \sum_{c=1}^{C} \sum_{w=1}^{W} \sum_{h=1}^{H} \lVert G(I) - \widetilde{J} \lVert
\end{equation}
Here, $\widetilde{J}$ represents $J$ in CIELAB color space. The difference between $G(I)$ and $J$ is fundamentally the difference of their $ab$ values.

The final objective function of the proposed generator network is

\begin{equation}
\label{eq:propsedganobjective}
    \Lagr_{CuPGAN} = \Lagr_{adv} + \lambda_{1} \Lagr_{L1} + \lambda_{2} \Lagr_{class} + \lambda_{3}  \Lagr_{per}
\end{equation}

Here $\Lagr_{adv}$ is the adversarial loss, $\Lagr_{L1}$ is the content-based per-pixel loss, $\Lagr_{class}$ is the classification loss and $\Lagr_{per}$ is the perceptual loss and $\lambda_{1},\lambda_{2},\lambda_{3}$ are positive constants. 

\subsection{Generator Network}
\label{sub:gennet}

\begin{figure}[!htb]
    \centering
    \includegraphics[width=\linewidth]{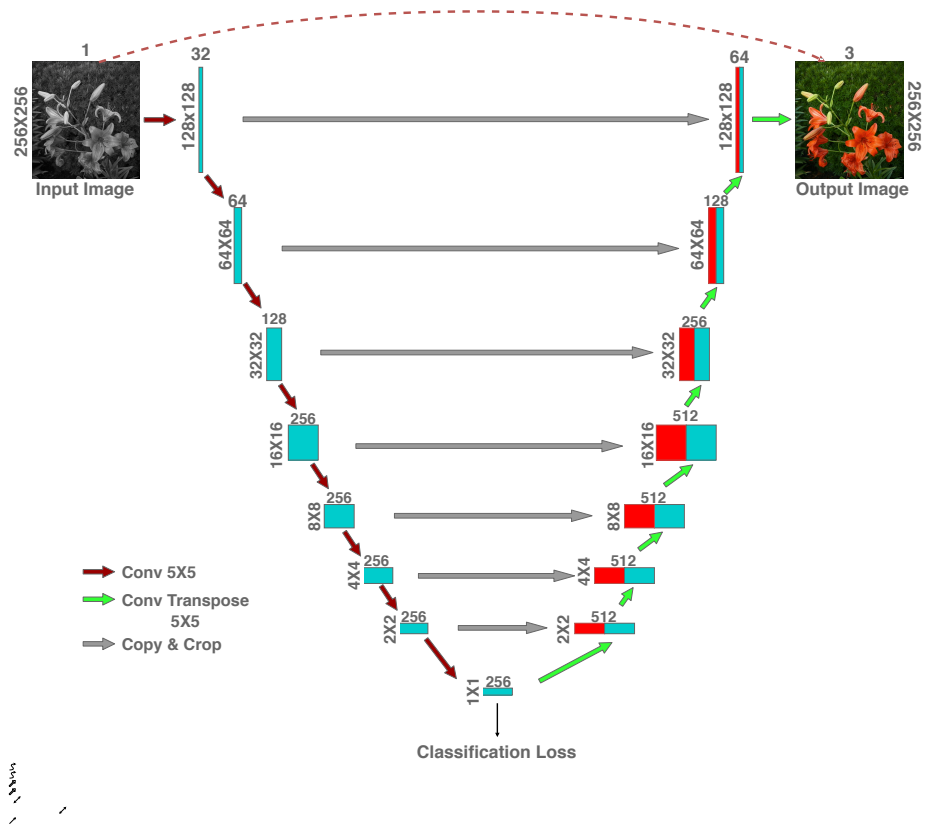}
    \caption{Network architecture of the proposed Generator network.}
    \label{fig:gennetwork}
\end{figure}

The generator network in the proposed architecture follows a similar architecture like U-Net by \cite{unet2015}. The U-Net was originally proposed for bio-medical image segmentation. The architecture of U-Net consists of a contracting path which acts as an encoder and expanding path which acts as a decoder. The skip connections are present to avoid any information and content loss due to convolutions. The success of U-Net propelled the utilization of U-Net architecture in many subsequent works. The elaborate generator network along with the feature map dimensions and the convolutional kernel sizes is shown in Fig.\ref{fig:gennetwork}. In the encoding process, the input feature map is reduced in height and width by a factor of 2 at every level. In the decoding process, the feature maps are enlarged in height and width by a factor of 2. At every decoding level, there is a step of feature-fusion from the opposite contracting path. The final transposed convolutional layer obtained are the $ab$ values which when concatenated with the $L$ (luminance) channel, gives us a colourized version of the input gray-scale image. The feature vector at the point of deflection of the encoding and the decoding path is processed through two fully-connected layers to obtain the vector for the classification loss. All the layers use the ReLU activation function except for the last transpose convolutional layer which uses the hyperbolic tangent (tanh) function. 

\subsection{Discriminator Network}
\label{sub:discrimnet}
\begin{figure}[!htb]
    \centering
    \includegraphics[width=\linewidth]{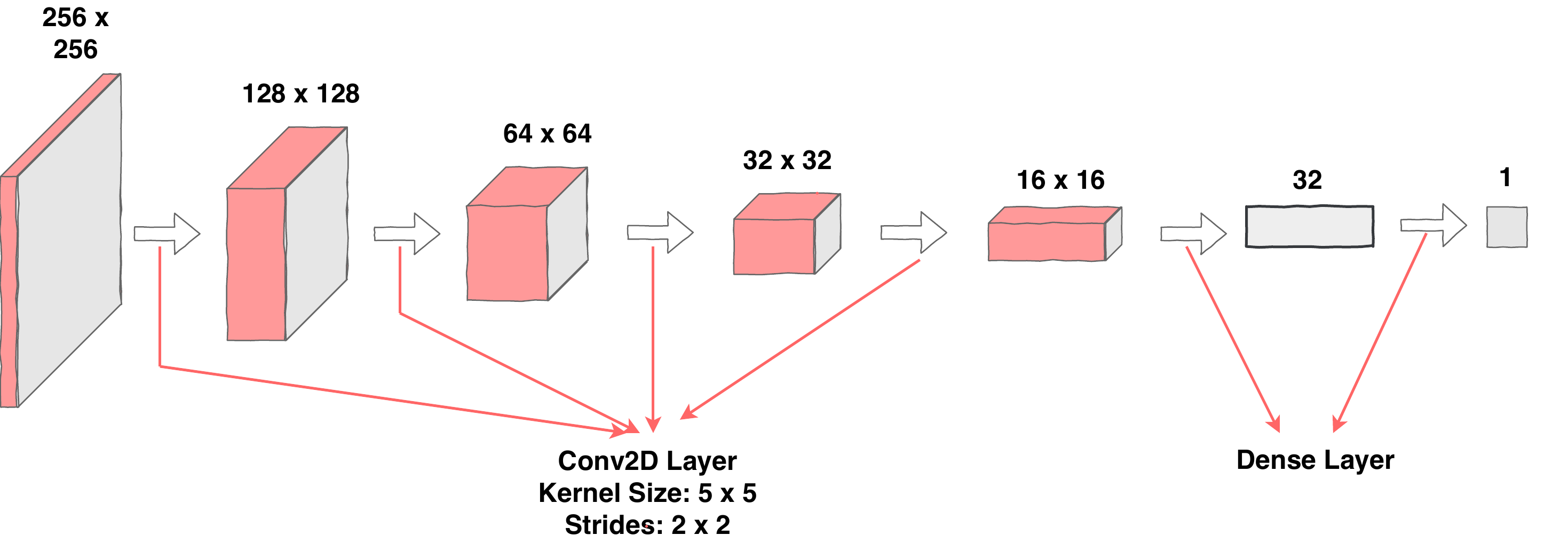}
    \caption{Network architecture of our Discriminator Network}
    \label{fig:discrimnetwork}
\end{figure}

The discriminator network used for the proposed method consists of four convolutional layers with kernel size $5 \times 5$ and stride length $2 \times 2$ followed by 2 fully-connected (FC) layers. The final FC-layer uses a Sigmoid activation to classify the image as real or fake. Fig.\ref{fig:discrimnetwork} shows the architecture of the discriminator network used for the proposed CuPGAN.  

\section{Experimental Results}
\label{sec:results}
\subsection{Experimental Settings}
All the training images are converted from RGB into the CIELAB colour space. Only the $L$ channel of images are fed into the network and the network tries to estimate the chromatic components $ab$. The values of the luminance and the chrominance components are normalized between $[-1, 1]$. The training images are re-sized to $256 \times 256$ in all cases. Every convolutional transpose layer in the decoder part of the generator network has a dropout with probability 0.5. Each layer in both the discriminator and the generator is followed by a batch-normalization layer. We use the Places365-Standard dataset \cite{zhou2017places} which contains about 1.8 million images. We filter the images that are grayscale and also those that have negligible colour variations which reduces our training dataset to 1.6 million images.  For our experiments, we empirically set $\lambda_{1} = 100$, $\lambda_{2} = 10$ and $\lambda_{3} = 1$ as the weights of the per-pixel, classification and perceptual loss respectively. We use the \textit{relu1\_2} layer of the VGGNet for computing the perceptual loss. The proposed system is implemented in Tensorflow. We train our network using the Adagrad optimizer with a learning rate of $10^{-4}$ and a batch-size of 16. Under these settings, the model is trained for 20 epochs on a system with NVIDIA TitanX Graphics Processing unit (GPU). 

\subsection{Quantitative Evaluations}
\label{sub:quaneval}

We evaluate our proposed algorithm on different metrics and compare it with the existing state-of-the art colourization methods like \cite{zhang2016colorful} and \cite{iizuka2016}. The metrics used for comparison are Peak Signal to Noise Ratio (PSNR) \cite{psnr}, Structural Similarity Index Measure (SSIM) \cite{ssim}, Mean Squared Error (MSE), Universal Quality Index (UQI) \cite{uqi} and Visual Information Fidelity (VIF) \cite{vif}. We carry out our quantitative comparison on the test dataset of Places365. It contains multiple images of the 365 different classes used in the training dataset. The quantitative evaluations are shown in Table \ref{tab:psnrplaces}. The proposed method CuPGAN performs better in all metrics except for the VIF and UQI where it shows competitive performance on the Places365 test dataset by \cite{zhou2017places}.

\begin{figure}[!htp]
    \centering
    \begin{tabular}{ccccc}
        \includegraphics[width=0.19\linewidth]{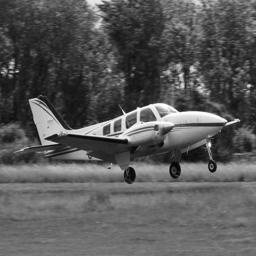}&\includegraphics[width=0.19\linewidth]{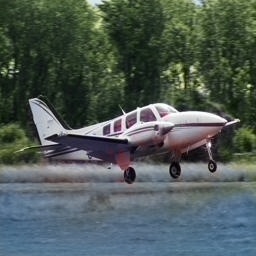}&\includegraphics[width=0.19\linewidth]{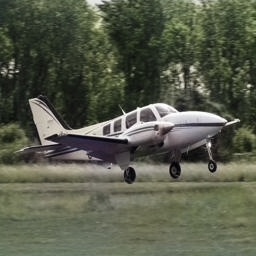}&\includegraphics[width=0.19\linewidth]{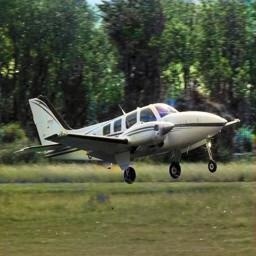}&\includegraphics[width=0.19\linewidth]{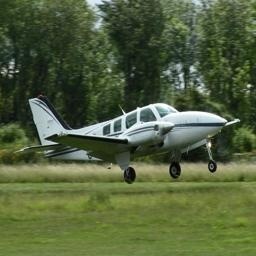} \\
        \includegraphics[width=0.19\linewidth]{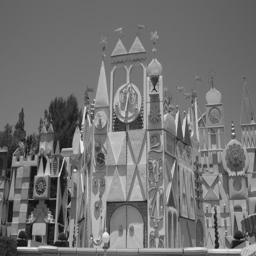}&\includegraphics[width=0.19\linewidth]{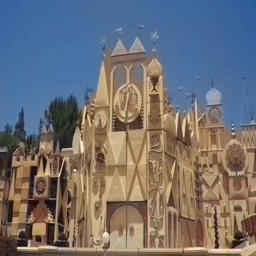}&\includegraphics[width=0.19\linewidth]{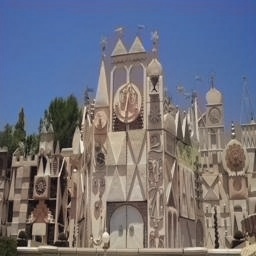}&\includegraphics[width=0.19\linewidth]{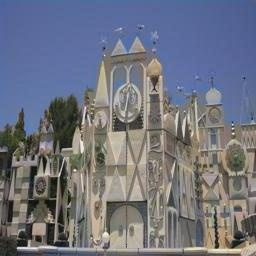}&\includegraphics[width=0.19\linewidth]{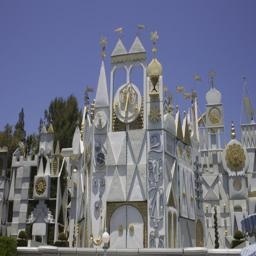} \\
        \includegraphics[width=0.19\linewidth]{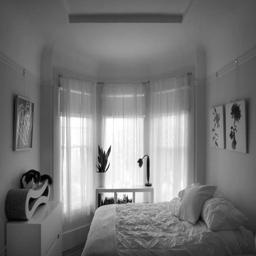}&\includegraphics[width=0.19\linewidth]{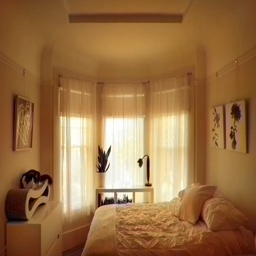}&\includegraphics[width=0.19\linewidth]{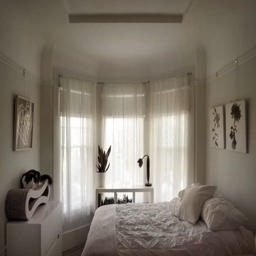}&\includegraphics[width=0.19\linewidth]{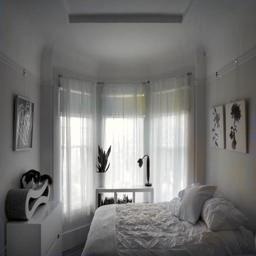}&\includegraphics[width=0.19\linewidth]{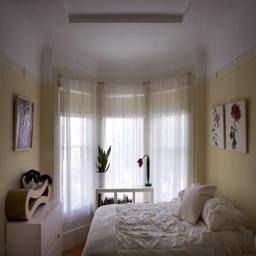} \\
        \includegraphics[width=0.19\linewidth]{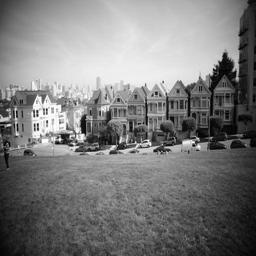}&\includegraphics[width=0.19\linewidth]{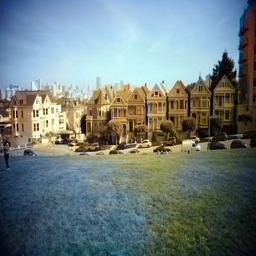}&\includegraphics[width=0.19\linewidth]{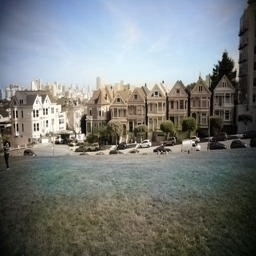}&\includegraphics[width=0.19\linewidth]{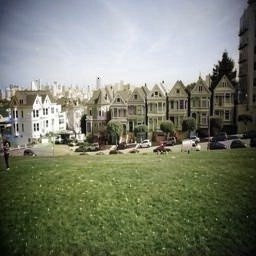}&\includegraphics[width=0.19\linewidth]{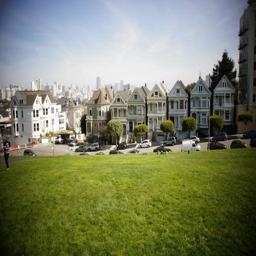} \\
        \includegraphics[width=0.19\linewidth]{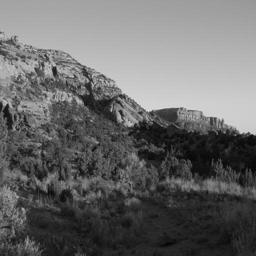}&\includegraphics[width=0.19\linewidth]{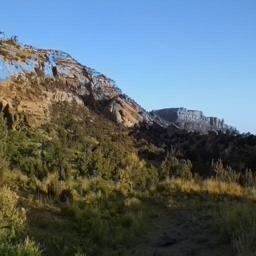}&\includegraphics[width=0.19\linewidth]{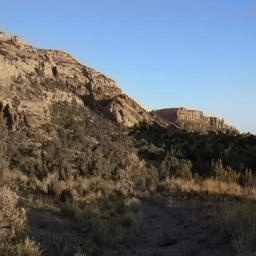}&\includegraphics[width=0.19\linewidth]{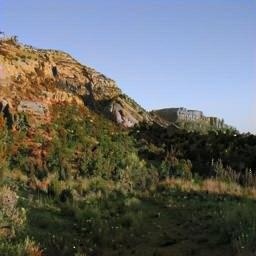}&\includegraphics[width=0.19\linewidth]{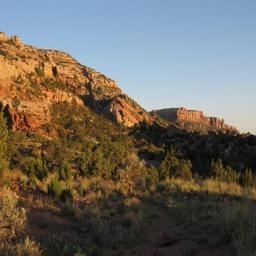} \\
        \includegraphics[width=0.19\linewidth]{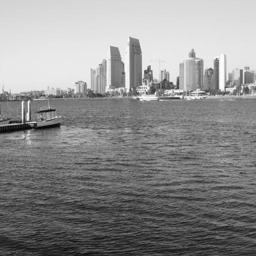}&\includegraphics[width=0.19\linewidth]{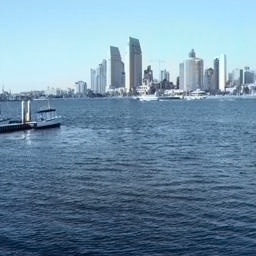}&\includegraphics[width=0.19\linewidth]{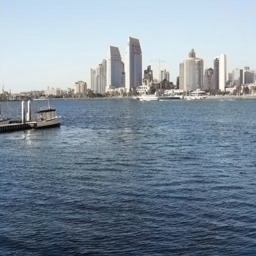}&\includegraphics[width=0.19\linewidth]{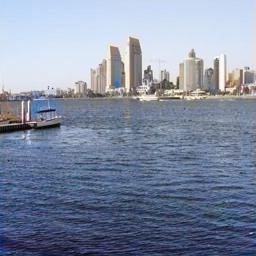}&\includegraphics[width=0.19\linewidth]{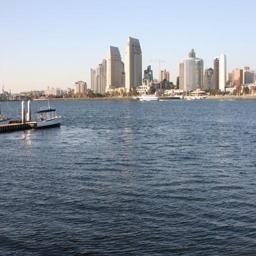} \\
        \includegraphics[width=0.19\linewidth]{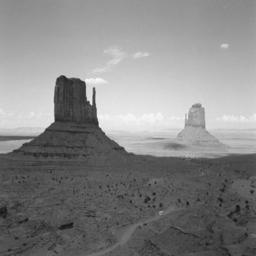}&\includegraphics[width=0.19\linewidth]{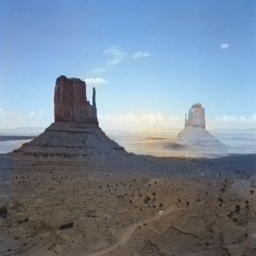}&\includegraphics[width=0.19\linewidth]{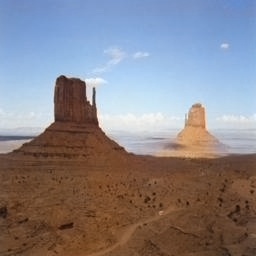}&\includegraphics[width=0.19\linewidth]{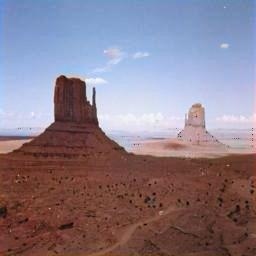}&\includegraphics[width=0.19\linewidth]{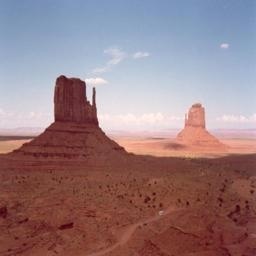} \\
        \includegraphics[width=0.19\linewidth]{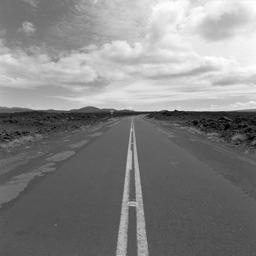}&\includegraphics[width=0.19\linewidth]{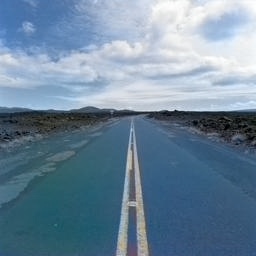}&\includegraphics[width=0.19\linewidth]{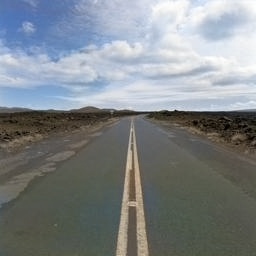}&\includegraphics[width=0.19\linewidth]{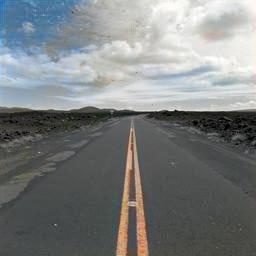}&\includegraphics[width=0.19\linewidth]{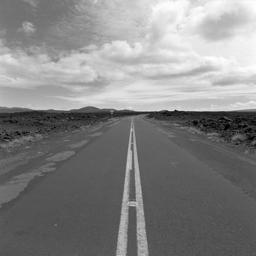} \\
        
        \\ Grayscale & Zhang \textit{et al.} & Iizuka \textit{et al.} & CuPGAN & Ground Truth 
    \end{tabular}
    \caption{Qualitative Comparison on the test partition of Places365 dataset \cite{zhou2017places}. The results display the robustness and versatility of our proposed method for colourizing both indoor and outdoor images}
    \label{fig:qual}
\end{figure}

\begin{table}[!htbp]
\centering
\setlength{\tabcolsep}{8pt}
\caption{Comparative Evaluation of PSNR, SSIM, MSE, UQI and VIF on Places365 dataset \cite{zhou2017places}}
\label{tab:psnrplaces}
\begin{tabular}{l l l l} \hline
\textbf{Method} & \cite{zhang2016colorful} & \cite{iizuka2016} & \textbf{CuPGAN} \\ \hline
PSNR & 25.44 & 27.14 & \textit{\textbf{28.41}} \\
SSIM & 0.95  & 0.95 & \textbf{\textit{0.96}} \\
MSE & 262.08 & 135.89 & \textbf{\textit{107.93}} \\
UQI & 0.56 & \textit{\textbf{0.60}} & 0.58\\
VIF & 0.886 & \textit{\textbf{0.905}} & 0.896 \\ \hline
\end{tabular}
\end{table}

\subsection{Qualitative Evaluations}
\label{sub:qualeval}

We demonstrate and compare our results in a qualitative manner against the existing methods \cite{iizuka2016} and \cite{zhang2016colorful}. Fig. \ref{fig:qual} displays the comparative qualitative results. We observe that in the first image \cite{zhang2016colorful} hallucinates bluish colour for the grass while \cite{iizuka2016} produces a less saturated image compared to ours and the ground-truth. In the second image, both methods \cite{iizuka2016} and \cite{zhang2016colorful}  tend to over-saturate the colour of the structure while our result is closest to the ground-truth. In the third image of the indoor scene, \cite{zhang2016colorful} tends to impart yellowish colour to the indoors contrasting our and \cite{zhang2016colorful}  result and also the ground-truth. In the fourth and the last image, both methods \cite{iizuka2016} and \cite{zhang2016colorful} tend to distort colours in the grass and the road respectively. Our method does not demonstrate any such colour distortions. In the fifth image, our method tends to over-estimate the grass regions near the foothills of the mountain but displays more vibrancy as compared to \cite{iizuka2016} and \cite{zhang2016colorful}. The results of the sixth and the seventh images are satisfactory for all methods. From what we observed, \cite{zhang2016colorful} tends to over-saturate the colours of the image. The qualitative comparison ensures that not only does our method produce images of crisp quality and of adequate saturation, but it also tends to produce less colour distortions and colour anomalies.

\subsection{Real World Images}
\label{sub:realworld}
\begin{figure}[!htp]
    \centering
    \begin{tabular}{cccc}
        \includegraphics[width=0.22\linewidth]{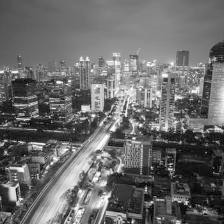}&\includegraphics[width=0.22\linewidth]{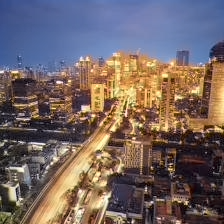}&\includegraphics[width=0.22\linewidth]{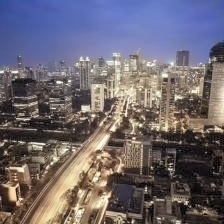}&\includegraphics[width=0.22\linewidth]{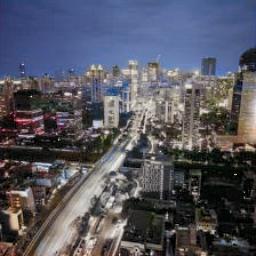}\\
        
        \includegraphics[width=0.22\linewidth]{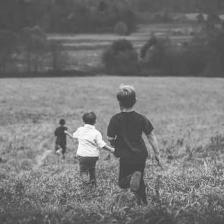}&\includegraphics[width=0.22\linewidth]{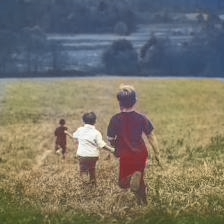}&\includegraphics[width=0.22\linewidth]{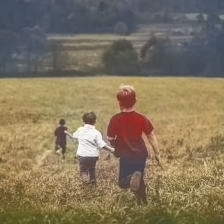}&\includegraphics[width=0.22\linewidth]{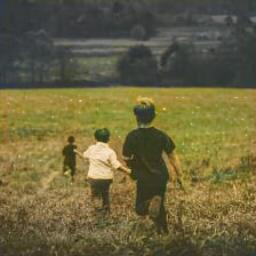}\\
        
        \includegraphics[width=0.22\linewidth]{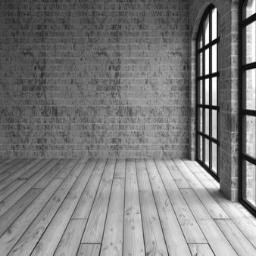}&\includegraphics[width=0.22\linewidth]{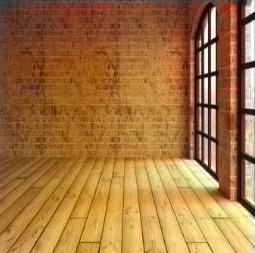}&\includegraphics[width=0.22\linewidth]{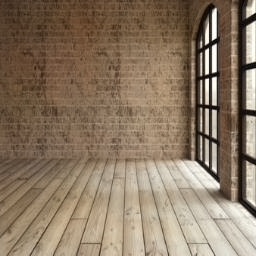}&\includegraphics[width=0.22\linewidth]{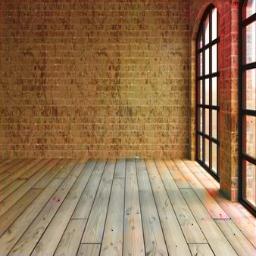}\\
        
        \includegraphics[width=0.22\linewidth]{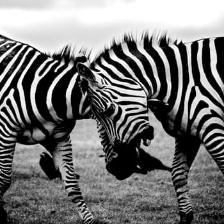}&\includegraphics[width=0.22\linewidth]{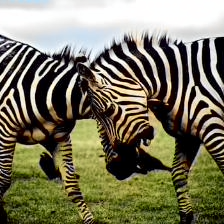}&\includegraphics[width=0.22\linewidth]{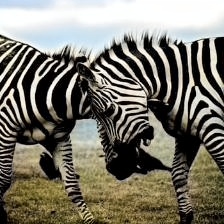}&\includegraphics[width=0.22\linewidth]{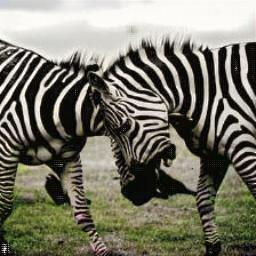}\\
        \includegraphics[width=0.22\linewidth]{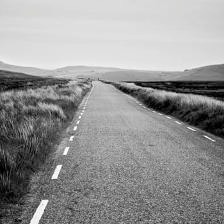}&\includegraphics[width=0.22\linewidth]{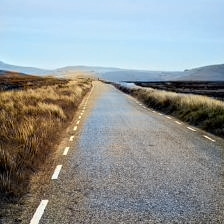}&\includegraphics[width=0.22\linewidth]{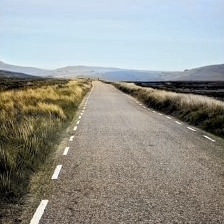}&\includegraphics[width=0.22\linewidth]{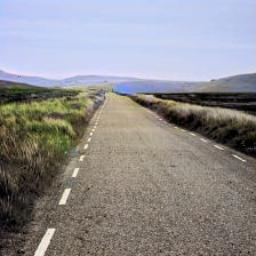}\\
        Grayscale & Zhang \textit{et al.}& Iizuka \textit{et al.}& CuPGAN
    \end{tabular}
    \caption{Qualitative Comparison on the random images from the internet}
    \label{fig:real}
\end{figure}

 In order to establish the efficiency of our model we collect some images from the internet randomly and colourize these images using all the competing methods used in the quantitative evaluations as discussed in Section \ref{sub:quaneval}. Fig. \ref{fig:real} shows the colourization on real world images. We can observe that \cite{zhang2016colorful} tends to over-saturate the colour components as visible from the first and the third image. Also, \cite{zhang2016colorful} produces some colour anomalies visible from the zebras (fourth image) and the road (fifth image). \cite{iizuka2016} tends to under-saturate the images as visible from the third and fourth image. The colourization from our proposed method produces more visually-appealing and sharp results comparatively.

\begin{figure}[!htp]
    \centering
    \begin{tabular}{ccccc}
        \includegraphics[width=0.19\linewidth]{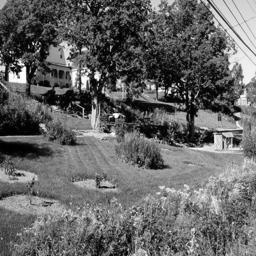}&\includegraphics[width=0.19\linewidth]{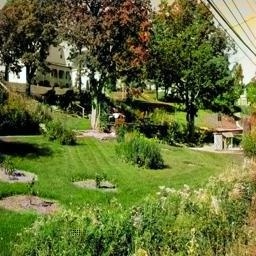}&\includegraphics[width=0.19\linewidth]{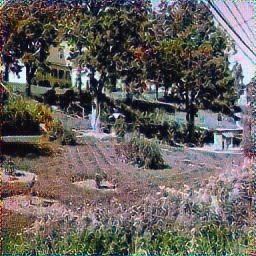}&\includegraphics[width=0.19\linewidth]{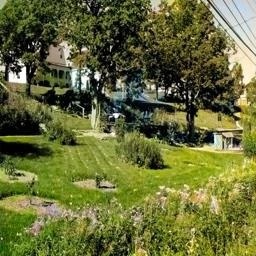}&\includegraphics[width=0.19\linewidth]{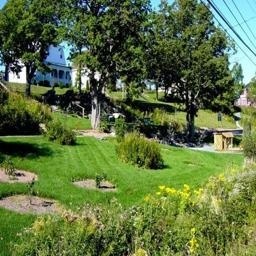} \\
        \includegraphics[width=0.19\linewidth]{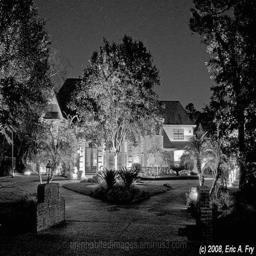}&\includegraphics[width=0.19\linewidth]{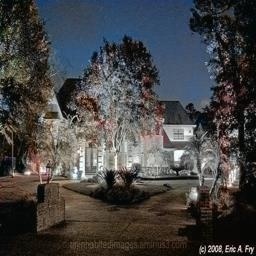}&\includegraphics[width=0.19\linewidth]{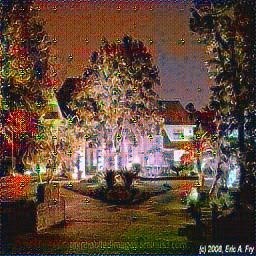}&\includegraphics[width=0.19\linewidth]{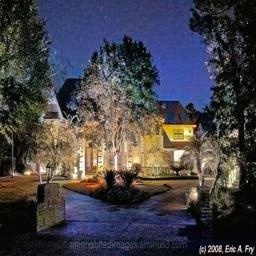}&\includegraphics[width=0.19\linewidth]{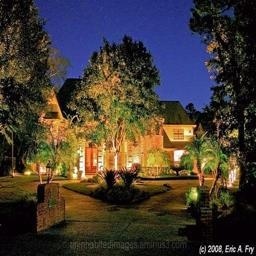} \\\includegraphics[width=0.19\linewidth]{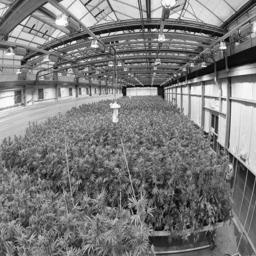}&\includegraphics[width=0.19\linewidth]{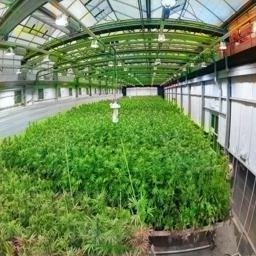}&\includegraphics[width=0.19\linewidth]{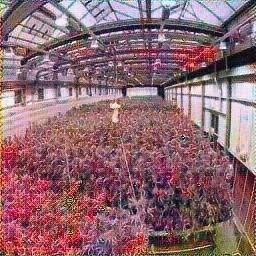}&\includegraphics[width=0.19\linewidth]{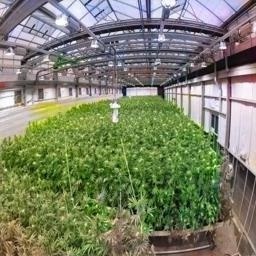}&\includegraphics[width=0.19\linewidth]{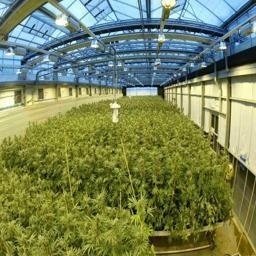} \\\includegraphics[width=0.19\linewidth]{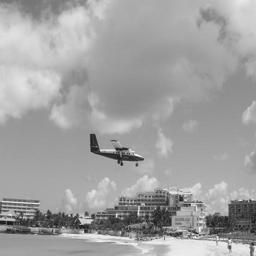}&\includegraphics[width=0.19\linewidth]{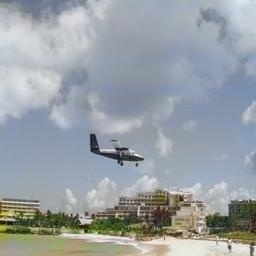}&\includegraphics[width=0.19\linewidth]{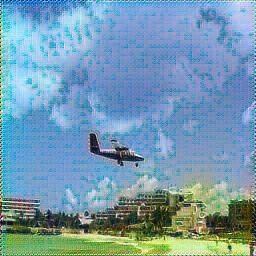}&\includegraphics[width=0.19\linewidth]{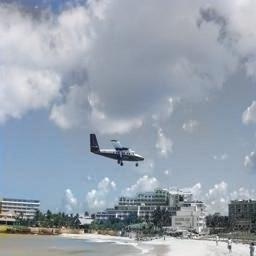}&\includegraphics[width=0.19\linewidth]{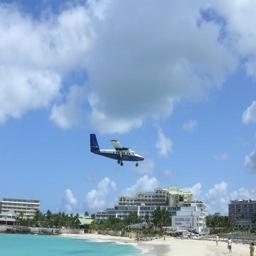} \\
        
        \\ Grayscale & $\Lagr_{L1}$ & $\Lagr_{per}$ & $\Lagr_{L1}+\Lagr_{per}$ (ours) & Ground Truth 
    \end{tabular}
    \caption{\textbf{Ablation}: Qualitative Comparison on the subset-test partition of Places365 dataset \cite{zhou2017places}}
    \label{fig:ablation}
\end{figure}

\subsection{Ablation studies}

For demonstrating the effectiveness of our loss function, we train our model using a subset of the Places365 data-set \cite{zhou2017places}. We created the subset by randomly selecting 30 different classes from the complete data-set. We train our proposed model using three different component loss functions: 1) using only $\Lagr_{L1}$ loss, 2) using only $\Lagr_{per}$ and 3) using both $\Lagr_{L1}$ and $\Lagr_{per}$ (\textit{ours}). The use of the first loss function corresponds directly to the objective function used in the \textit{cGAN} model proposed by \cite{isola2017}. We provide both qualitative and quantitative comparisons for validating our point. The qualitative results are displayed in \ref{fig:ablation}. The results display the effectiveness of using both $\Lagr_{L1}$ and $\Lagr_{per}$ loss together over using them discretely. Table \ref{tab:ablation} displays the quantitative comparison of using the mentioned loss functions. We can infer both qualitatively and quantitatively that our objective function performs better than using only $\Lagr_{L1}$ or only only $\Lagr_{per}$.

\begin{table}[!htbp]
\centering
\setlength{\tabcolsep}{8pt}
\caption{Evaluation of PSNR, SSIM, MSE, UQI and VIF on subset test data-set from Places365 \cite{zhou2017places}}
\label{tab:ablation}
\begin{tabular}[width=0.5\linewidth]{llll} \hline
\textbf{Method} & \textbf{$\Lagr_{L1}$}& \textbf{$\Lagr_{per}$} & \textbf{Ours}\\ \hline
PSNR & 22.00 & 17.40  & \textit{\textbf{23.19}} \\
SSIM & 0.88 & 0.63  & \textbf{\textit{0.91}} \\
MSE & 552.42 & 1341.66  & \textbf{\textit{529.53}} \\
UQI & \textbf{\textit{0.50}} & 0.20  & 0.49 \\
VIF & 0.85 & 0.53 & \textbf{\textit{0.86}} \\ \hline
\end{tabular}
\end{table}

\subsection{Historic black and white images}
\begin{figure}
 \centering
  \label{fig:historic}
 \begin{tabular}{cccc}
  \includegraphics[width=0.2\linewidth, height=1.2in]{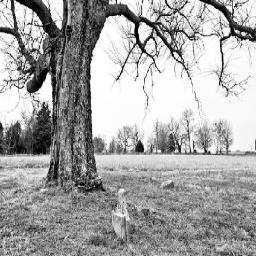} & \includegraphics[width=0.2\linewidth, height=1.2in]{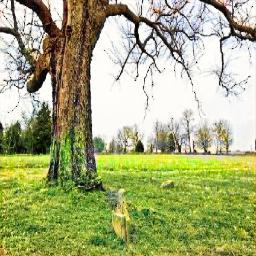}& 
  \includegraphics[width=0.2\linewidth, height=1.2in]{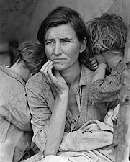} & \includegraphics[width=0.2\linewidth, height=1.2in]{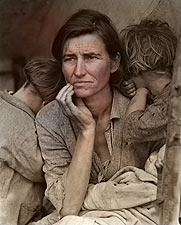}

 \end{tabular}
 \caption{Colourization of historic black and white images}
\end{figure}

We also test our model on $20^{th}$ century black and white images. Due to the type of films and cameras used in the past, we can never be perfectly sure about the type of colours and shades that was originally there. The images used have significant artifacts and have irregular borders which makes it an ill-posed task to reckon colours. In spite of all these issues, our model is able to colourize these images producing satisfactory results. In Fig. \ref{fig:historic} we show some examples of colourization of these historic images and we observe visually pleasing results.

\begin{figure}
 \centering
    \begin{tabular}{ccccc}
    \includegraphics[width=0.15\linewidth, height=0.8in]{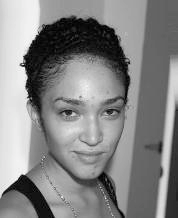}&
    \includegraphics[width=0.15\linewidth, height=0.8in]{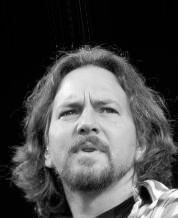} &
    \includegraphics[width=0.15\linewidth, height=0.8in]{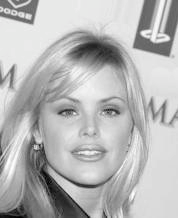} &
    \includegraphics[width=0.15\linewidth, height=0.8in]{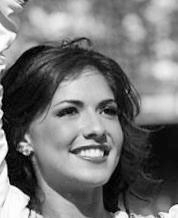} &
    \includegraphics[width=0.15\linewidth, height=0.8in]{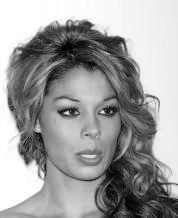} \\
    \includegraphics[width=0.15\linewidth, height=0.8in]{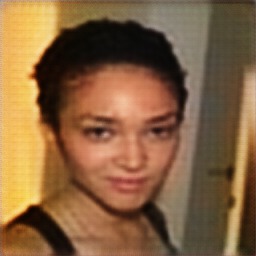}&
    \includegraphics[width=0.15\linewidth, height=0.8in]{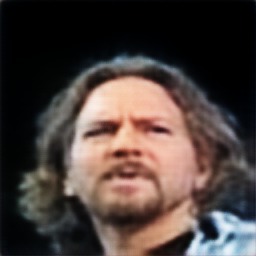} &
    \includegraphics[width=0.15\linewidth, height=0.8in]{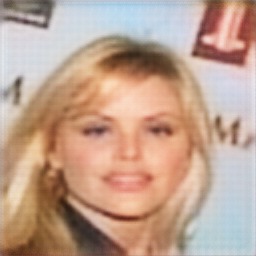} &
    \includegraphics[width=0.15\linewidth, height=0.8in]{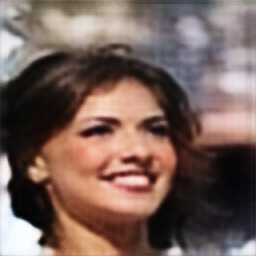} &
    \includegraphics[width=0.15\linewidth, height=0.8in]{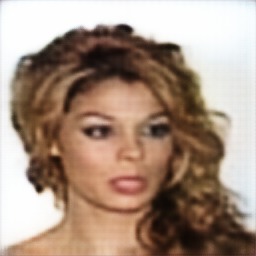} \\
    \includegraphics[width=0.15\linewidth, height=0.8in]{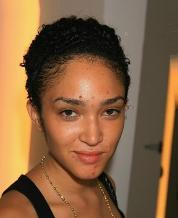} &
    \includegraphics[width=0.15\linewidth, height=0.8in]{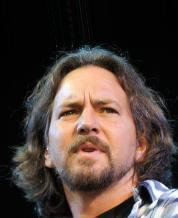} &
    \includegraphics[width=0.15\linewidth, height=0.8in]{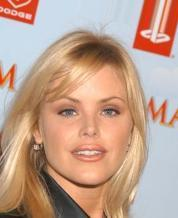} &
    \includegraphics[width=0.15\linewidth, height=0.8in]{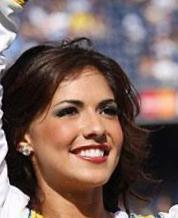} &
    \includegraphics[width=0.15\linewidth, height=0.8in]{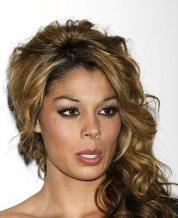} \\
   \end{tabular}
  \caption{Visual results on the CelebA dataset \cite{celeba}}
 \label{fig:celebqual}
\end{figure}

\subsection{Adaptability}
In order to establish the adaptability of our proposed method, we train our model on a very different dataset than the Places365 dataset \cite{zhou2017places}. We use the large-scale CelebFaces Attributes or the CelebA dataset by \cite{celeba} for training. CelebA is a large-scale dataset which contains more than 200K images with 40 attributes. The classification loss of the generator objective is calculated using the cross-entropy loss of the attributes. We provide a visual analysis and quantitative evaluation of our method on this dataset as well. We use the metrics described in \ref{sub:quaneval}. Table \ref{tab:placesquan} shows the evaluation of our model on the CelebA dataset. The quantitative evaluations assure that our model can be employed easily to another dataset for the process of colourization. Fig. \ref{fig:celebqual} demonstrates the visual results on the CelebA dataset by \cite{celeba} supplements our claim of adaptability to versatile datasets.

\begin{table}[!htp]
\centering
\caption{Quantitative evaluations on the CelebA dataset \cite{celeba}}
\label{tab:placesquan}
\begin{tabular}{p{4cm}   l}
\hline
Metrics     & CuPGAN \\
\hline
PSNR    & \textbf{25.9} \\
SSIM    & \textbf{0.89} \\
UQI     & \textbf{0.70} \\
VQI     & \textbf{0.78} \\
\hline
\end{tabular}
\end{table}

\section{Conclusion}
\label{sec:conclusion}
In this paper, we develop on the Conditional Generative Adversarial Networks (cGANs) framework to deal with the task of image colourization by incorporating the recently flourished perceptual loss and cross-entropy classification loss. We train our proposed model CuPGAN in an end-to-end fashion. Quantitative and qualitative evaluations establish the significant enhancing effects of adding the perceptual and classification loss as compared to the vanilla Conditional-GANs. Also, experiments conducted on standard data-sets show promising results when compared to the standard exclusive state-of-the art image colourization methods evaluated using five standard image quality measures like PSNR, SSIM, MSE, UIQ and VIF. Our proposed method performs appreciably well producing clear and crisp quality colourized pictures even in cases of images picked from the internet and historic black and white images. 

\section{Acknowledgements}
We would like to thank Mr. Ayan Kumar Bhunia for his guidance and help regarding implementation and development of the model. We would also like to thank Mr. Ayush Daruka and Mr. Prashant Kumar for their valuable discussions.  
\bibliographystyle{plainnat}
\bibliography{0955}
\end{document}